\pdfoutput=1
\documentclass[11pt]{article}

\usepackage{EMNLP2022}

\usepackage{times}
\usepackage{latexsym}
\usepackage{adjustbox}

\usepackage[T1]{fontenc}
\usepackage[utf8]{inputenc}

\usepackage{microtype}
\usepackage[ruled]{algorithm2e}
\usepackage{graphicx}
\usepackage{multirow}
\usepackage{latexsym}
\usepackage{amsmath,amssymb,amsthm}
\usepackage{subfigure}
\usepackage{microtype}
\DeclareGraphicsExtensions{.pdf, .png}
\usepackage{epstopdf}

\usepackage{booktabs}
\usepackage{setspace}
\usepackage[normalem]{ulem}
\useunder{\uline}{\ul}{}
\usepackage{tikz}
\usepackage{bbding}
\usepackage{pifont}
\usepackage{wasysym}
\usepackage{paralist}

\title{UniMSE: Towards Unified Multimodal Sentiment Analysis \\and Emotion Recognition}
\author{Guimin Hu$^{\dagger}$, Ting-En Lin, Yi Zhao$^{\dagger}$\thanks{\ \ Corresponding authors.}, Guangming Lu$^{\dagger}$, Yuchuan Wu, Yongbin Li$^{*}$  \\
  $^{\dagger}$School of Computer Science and Technology, School of Science, \\ Harbin Institute of Technology (Shenzhen), China \\
  \texttt{\{rice.hu.x,liyb821\}@gmail.com} \\ 
  \texttt{\{zhao.yi,luguangm\}@hit.edu.cn}}
\begin{document}
\maketitle
\begin{abstract}
%%% abstract突出的几个点：
% Sentiment knowledge in common exists in many emotion-related tasks, but the knowledge is isolated due to the difference of downstream tasks and datasets. In the fields of multimodal sentiment analysis (MSA) and emotion recognition in conversation (ERC), MSA aims to predict a real number that reflects the sentiment strength, and ERC aims to predict the emotion category of each utterance. We believe that MSA and ERC share a feature space and there exist mutual indication between them in multimodal representation learning. In this paper, we propose a unified multimodal sentiment knowledge sharing framework (UniMSE) towards MSA and ERC tasks, which breaks the information boundary and maximize the complementarity and consistency in sentiment expression between the two tasks. Besides, we propose to fuse multimodal representation from syntax and semantic levels and utilize the contrastive learning to differentiate the multimodal fusion representation among samples. We conduct experiments on four benchmark datasets of MSA and ERC, and achieve the state-of-the-art performance.

Multimodal sentiment analysis (MSA) and emotion recognition in conversation (ERC) are key research topics for computers to understand human behaviors. From a psychological perspective, emotions are the expression of affect or feelings during a short period, while sentiments are formed and held for a longer period. However, most existing works study sentiment and emotion separately and do not fully exploit the complementary knowledge behind the two. In this paper, we propose a multimodal sentiment knowledge-sharing framework (UniMSE) that unifies MSA and ERC tasks from features, labels, and models. We perform modality fusion at the syntactic and semantic levels and introduce contrastive learning between modalities and samples to better capture the difference and consistency between sentiments and emotions. Experiments on four public benchmark datasets, MOSI, MOSEI, MELD, and IEMOCAP, demonstrate the effectiveness of the proposed method and achieve consistent improvements compared with state-of-the-art methods.
\end{abstract}

\section{Introduction}

With the rapid development of multimodal machine learning \cite{liang2022foundations, baltruvsaitis2018multimodal} and dialog system \cite{he2022tree, he2022unified, he2022galaxy}, Multimodal Sentiment Analysis (MSA) and Emotion Recognition in Conversations (ERC) have become the keys for machines to perceive, recognize, and understand human behaviors and intents \cite{Zhang_Xu_Lin_2021, Zhang_Xu_Lin_Lyu_2021, DBLP:conf/emnlp/HuLZ21,DBLP:journals/kbs/HuLZ21}. Multimodal data provides not only verbal information, such as textual (spoken words) features but also non-verbal information, including acoustic (prosody, rhythm, pitch) and visual (facial attributes) features. These different modalities allow the machine to make decisions from different perspectives, thereby achieving more accurate predictions \cite{DBLP:conf/icml/NgiamKKNLN11}. The goal of MSA is to predict sentiment intensity or polarity, and ERC aims to predict predefined emotion categories. There are many research directions in MSA and ERC, such as multimodal fusion \cite{DBLP:conf/naacl/YangWYZRZPM21}, modal alignment \cite{DBLP:conf/acl/TsaiBLKMS19}, context modeling \cite{DBLP:conf/emnlp/Mao0WGL21} and external knowledge \cite{DBLP:conf/emnlp/GhosalMGMP20}. However, most existing works treat MSA and ERC as separate tasks, ignoring the similarities and complementarities between sentiments and emotions. 

\begin{figure}[t]
\centering
\includegraphics[width=0.95\linewidth]{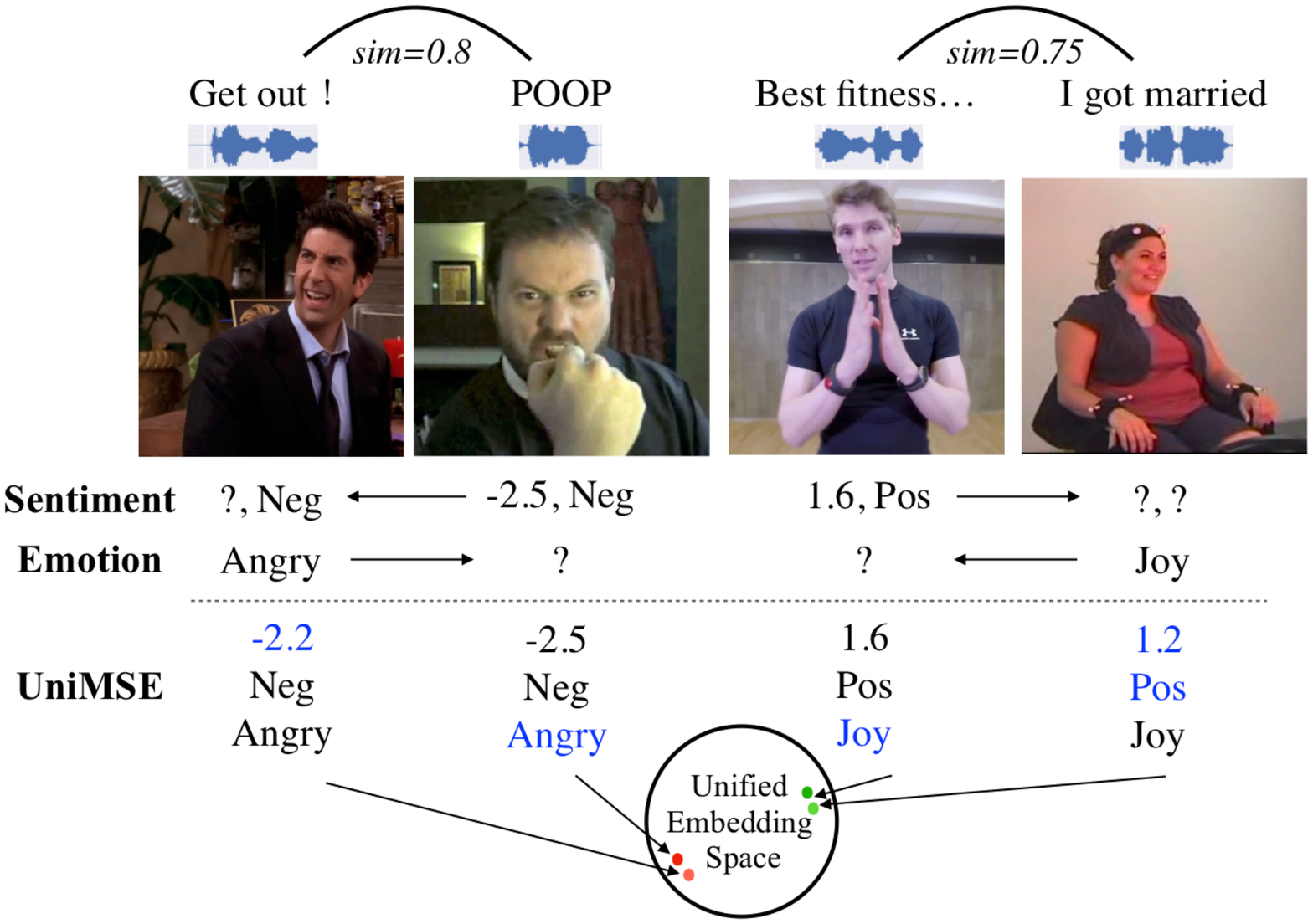}
\caption{Illustration of sentiment and emotion sharing a unified embedding space. The bottom is a unified label after formalizing sentiment and emotion according to the similarity $\overset{\frown}{sim}$ between samples with the same sentiment polarity label.}
\label{fig:example}
\end{figure}

On the one hand, from a psychological perspective, both sentiments and emotions are experiences that result from the combined influences of the biological, cognitive, and social \cite{stets2006emotions}, and could be expressed similarly. In Figure \ref{fig:example}, we illustrate how sentiments and emotions are relevant in the verbal or non-verbal, and could be projected into a unified embedding space. On the other hand, emotions are reflections of the perceived change in the present within a short period \cite{batson1992differentiating}, while sentiments are held and formed in longer periods \cite{murray1945clinical}. In our preliminary study, we found that the video duration of MSA is almost twice of ERC\footnote{Please see Appendix \ref{sec:duration} for details.}, which is consistent with the above definitions. A variety of psychological literature \cite{davidson2009handbook,ben2001subtlety,shelly2004emotions} explain the similarities and differences between sentiment and emotion. \citet{DBLP:journals/taffco/MunezeroMSP14} also investigates the relevance and complementarity between the two and point out that analyzing sentiment and emotion together could better understand human behaviors.

Based on the above motivation, we propose a multimodal sentiment knowledge-sharing framework that \textbf{Uni}fied \textbf{MS}A and \textbf{E}RC (UniMSE) tasks. UniMSE reformulates MSA and ERC as a generative task to unify input, output, and task. We extract and unify audio and video features and formalize MSA and ERC labels into Universal Labels (UL) to unify sentiment and emotion.

Besides, previous works on multimodal fusion at multi-level textual features \cite{DBLP:conf/naacl/PetersNIGCLZ18,DBLP:conf/nips/VaswaniSPUJGKP17}, like syntax and semantics, are lacking. Therefore, we propose a pre-trained modality fusion layer (PMF) and embed it in Transformer \cite{DBLP:conf/nips/VaswaniSPUJGKP17} layers of T5 \cite{DBLP:journals/jmlr/RaffelSRLNMZLL20}, which fuses the acoustic and visual information with different level textual features for probing richer information. Last but not least, we perform inter-modal contrastive learning (CL) to minimize intra-class variance and maximize inter-class variance across modalities.

Our contributions are summarized as follows:
\begin{compactitem}
\item[1.] We propose a multimodal sentiment-knowledge sharing framework\footnote{https://github.com/LeMei/UniMSE.} (UniMSE) that unifies MSA and ERC tasks. The proposed method exploits the similarities and complementaries between sentiments and emotions for better prediction.

\item[2.] We fuse multimodal representation from multi-level textual information by injecting acoustic and visual signals into the T5 model. Meanwhile, we utilize inter-modality contrastive learning to obtain discriminative multimodal representations.

\item[3.] Experimental results demonstrate that UniMSE achieves a new state-of-the-art performance on four public benchmark datasets, MOSI, MOSEI, MELD and IEMOCAP, for MSA and ERC tasks. 

\item[4.] To the best of our knowledge, we are the first to solve MSA and ERC in a generative fashion, and the first to use unified audio and video features across MSA and ERC tasks. %\footnote{We believe the release of unified features is valuable to the research community.}
\end{compactitem}

% : 1) multimodal fusion, 2) multimodal consistency and difference, 3) multimodal alignment, and 4) multimodal context
\section{Related Work}
\paragraph{Multimodal Sentiment Analysis (MSA)} 
MSA aims to predict sentiment polarity and sentiment intensity under a multimodal setting \cite{DBLP:conf/icmi/MorencyMD11}. MSA research could be divided into four groups. The first is multimodal fusion. Early works of multimodal fusion mainly operate geometric manipulation in the feature spaces \cite{DBLP:conf/emnlp/ZadehCPCM17}. The recent works develop the reconstruction loss \cite{DBLP:conf/mm/HazarikaZP20}, or hierarchical mutual information maximization \cite{DBLP:conf/emnlp/HanCP21} to optimize multimodal representation. The second group focuses on modal consistency and difference through multi-task joint learning \cite{DBLP:conf/aaai/YuXYW21} or translating from one modality to another \cite{DBLP:conf/aaai/Mai0X20}. 
The third is multimodal alignment. \citet{DBLP:conf/acl/TsaiBLKMS19} and \citet{DBLP:journals/corr/abs-2112-01368} leverage cross-modality and multi-scale modality representation to implement modal alignment, respectively. Lastly, studies of multimodal context integrate the unimodal context, in which \citet{DBLP:conf/emnlp/ChauhanAEB19} adapts context-aware attention, \citet{DBLP:conf/emnlp/GhosalACPEB18} uses multi-modal attention, and \citet{DBLP:conf/icdm/PoriaCHMZM17} proposes a recurrent model with multi-level multiple attentions to capture contextual information among utterances. 
% or the loss with multi-scale fusion . Specially, \citet{DBLP:conf/aaai/YuXYW21} designed a label generation module to acquire independent unimodal supervisions and jointly trained the multi-modal and uni-modal tasks. \citet{DBLP:conf/acl/TsaiBLKMS19} modeled the interactions between modalities by directional pairwise cross-modal Transformer \cite{DBLP:conf/nips/VaswaniSPUJGKP17}. 
% The development of social media, such as movies, short-form videos, bring rich heterogeneous resources, so as to give a boost to MSA.

\begin{figure*}[t]
\centerline{\includegraphics[width=0.94\textwidth]{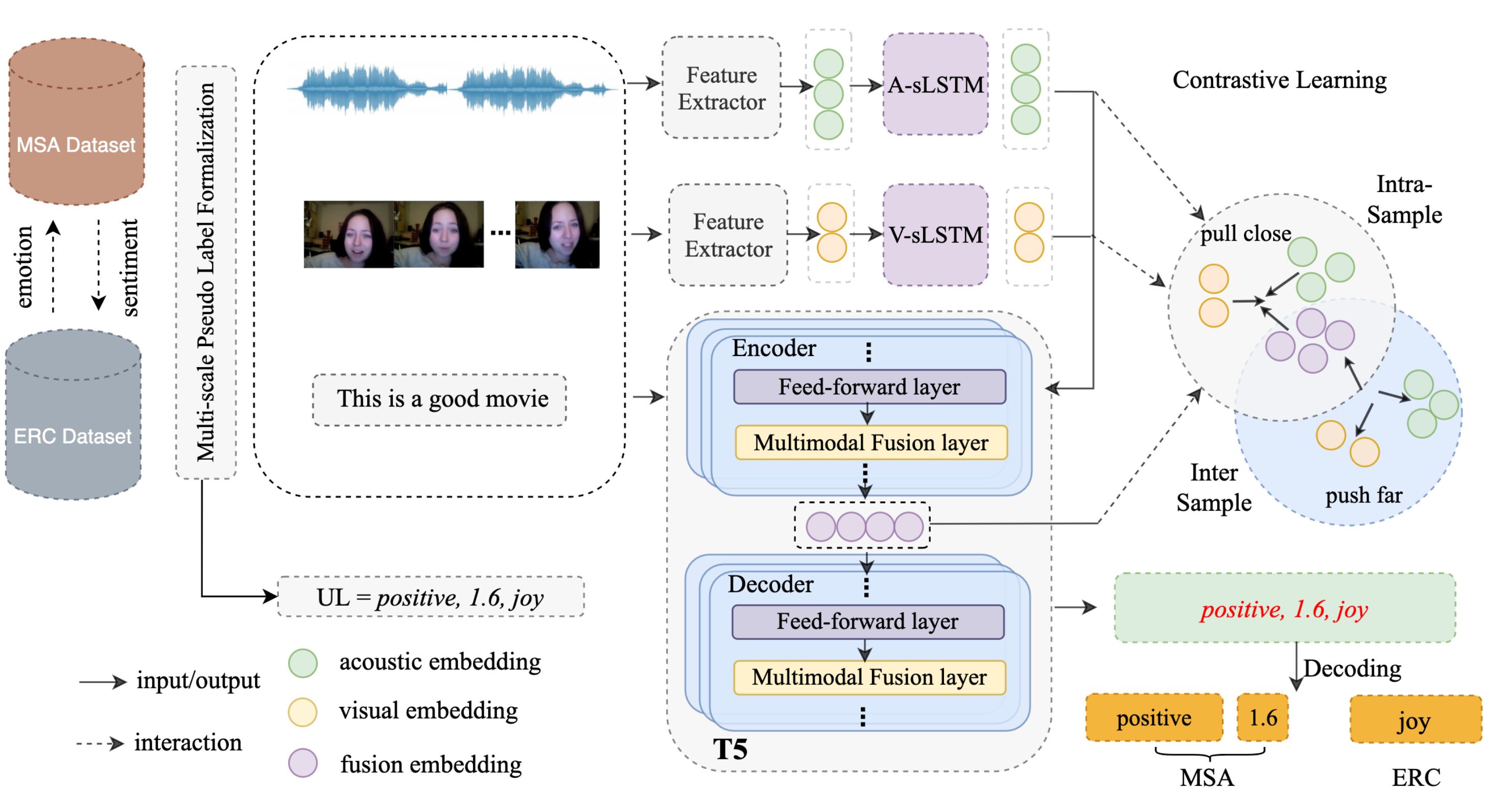}}
\caption{The overview of UniMSE.}
\label{fig:architecture}
\end{figure*}

\paragraph{Emotion Recognition in Conversations (ERC)}
With growing research interest in dialogue systems \cite{dai2021preview, dai2020learning, dai2020survey, lin2019deep, lin2019post, lin2020discovering, zhang-etal-2022-slot}, how to recognize the predefined emotion in the conversation has become a research hotspot. Meanwhile, with the rise of multimodal machine learning \cite{mao2022m, yuan2021transformer, yu2021learning, zhang2022mintrec, lin2022duplex}, the studies of ERC have been extended to multimodal paradigm. The multimodal emotion recognition in conversation gained great progress. The research direction could be categorized into multimodal fusion, context-aware models, and incorporating external knowledge. \citet{DBLP:conf/icassp/HuHWJM22, DBLP:conf/acl/HuLZJ20, DBLP:journals/corr/abs-2205-02455} adopt graph neural networks to model the inter/intra dependencies of utterances or speakers. For context incorporation, \citet{DBLP:conf/emnlp/SunYF21,DBLP:conf/emnlp/Li00W21,DBLP:conf/emnlp/GhosalMPCG19} model the contexts by constructing graph structure, and \citet{DBLP:conf/emnlp/Mao0WGL21} introduces the concept of emotion dynamics to capture context. Moreover, some advancing ERC works incorporate external knowledge, such as transfer learning \cite{DBLP:journals/corr/abs-1910-04980,DBLP:journals/corr/abs-2108-11626}, commonsense knowledge \cite{DBLP:conf/emnlp/GhosalMGMP20}, multi-task learning \cite{DBLP:conf/naacl/AkhtarCGPEB19}, and external information \cite{DBLP:conf/acl/ZhuP0ZH20} to solve ERC task.

% scLSTM \cite{DBLP:conf/acl/PoriaCHMZM17} captured the self-dependency through a bidirectional LSTM. CMN \cite{DBLP:conf/naacl/HazarikaPZCMZ18} and ICON \cite{DBLP:conf/emnlp/HazarikaPMCZ18} leveraged memory network to model separately the self and inter-personal dependencies. DialogueRNN \cite{DBLP:conf/aaai/MajumderPHMGC19} further developed ERC to multi-party conversations. DialogueGCN \cite{DBLP:conf/emnlp/GhosalMPCG19} and DAG-ERC \cite{DBLP:conf/acl/ShenWYQ20} build the graph structure and perform Graph Neutral Network (GNN) on it to solve ERC task. BiERU \cite{DBLP:journals/ijon/LiSJC22} aims to model the party-ignorant transferring of emotion in a conversation. EmotionFlow \cite{DBLP:conf/icassp/SongZZHH22} incorporates the participants’ emotions and DialogueTRM-M \cite{DBLP:conf/emnlp/Mao0WGL21} extends the concept of emotion dynamics. More recently, some advancing ERC works, such as transfer learning \cite{DBLP:journals/corr/abs-1910-04980,DBLP:journals/corr/abs-2108-11626} commonsense knowledge \cite{DBLP:conf/emnlp/GhosalMGMP20}, external knowledge \cite{DBLP:conf/acl/ZhuP0ZH20}, speaker and contextual incorporation \cite{DBLP:journals/corr/abs-2108-12009,DBLP:journals/corr/abs-2108-11626}, and curriculum learning \cite{DBLP:journals/corr/abs-2112-11718}, have employed pre-training models to solve ERC task. 

\paragraph{Unified Framework}
In recent years, the unification of related but different tasks into a framework has achieved significant progress \cite{DBLP:journals/corr/abs-2204-04637,DBLP:journals/corr/abs-2201-05966,DBLP:conf/aaai/ZhangMWJLY22}. For example, T5 \cite{DBLP:journals/jmlr/RaffelSRLNMZLL20} unifies various NLP tasks by casting all text-based language problems as a text-to-text format and achieves state-of-the-art results on many benchmarks. More recently, the works \cite{DBLP:journals/corr/abs-2111-02358,DBLP:conf/emnlp/ChengYZS21,DBLP:journals/corr/abs-2111-02358} using unified frameworks have attracted lots of attention, such as \citet{DBLP:conf/acl/YanDJQ020} solves all ABSA tasks in a unified index generative way,  \citet{DBLP:journals/corr/abs-2204-04637} investigates a unified generative dialogue understanding framework, \citet{DBLP:journals/corr/abs-2109-05812} proposes a unified framework for multimodal summarization, \citet{DBLP:conf/acl/WangSWZLY20} unifies entity detection and relation classification on their label space to eliminate the different treatment, and \citet{DBLP:conf/acl/YanGDGZQ20} integrates the flat NER, nested NER, and discontinuous NER subtasks in a Seq2Seq framework. These works demonstrate the superiority of such a unified framework in improving model performance and generalization. In our work, we use T5 as the backbone to unify the MSA and ERC and learn a unified embedding space in this framework.

\section{Method}

\subsection{Overall Architecture}
%%% overall architecture 需要写到模型输入、模型输出(多尺度的预测标签到MSA和ERC标签的映射)、融合层、对比学习。
As shown in Figure \ref{fig:architecture}, UniMSE comprises the task formalization, pre-trained modality fusion and inter-modality contrastive learning. First, we process off-line the labels of MSA and ERC tasks into the universal label (UL) format. Then we extract separately audio and video features using unified feature extractors among datasets. After obtaining audio and video features, we feed them into two individual LSTMs to exploit long-term contextual information. For textual modality, we use the T5 as the encoder to learn contextual information on the sequence. Unlike previous works, we embed multimodal fusion layers into T5, which follows the feed-forward layer in each of several Transformer layers of T5. Besides, we perform inter-modal contrastive learning to differentiate the multimodal fusion representations among samples. Specifically, contrastive learning aims to narrow the gap between modalities of the same sample and push the modality representations of different samples further apart.

\subsection{Task Formalization}
Given a multimodal signal $I_{i}=\{I^{t}_{i}, I^{a}_{i}, I^{v}_{i}\}$, we use $I^{m}_{i}, m\in \{t, a, v\}$ to represent unimodal raw sequence drawn from the video fragment $i$, where $\{t, a, v\}$ denote the three types of modalities—text, acoustic and visual. MSA aims to predict the real number $y^{r}_{i}\in \mathbb{R}$ that reflects the sentiment strength, and ERC aims to predict the emotion category of each utterance. MSA and ERC are unified in input feature, model architecture, and label space through task formalization. Task formalization contains input formalization and label formalization, where input formalization is used to process the dialogue text and modal feature, and label formalization is used to unify MSA and ERC tasks by transferring their labels into universal labels. Furthermore, we formalize the MSA and ERC as a generative task to unify them in a single architecture.

\subsubsection{Input Formalization}
%%% 这部分我主要写对于ERC任务在输入上的处理，拼接前后两轮的utter并用分割符分割，采用segment id进行区分。
The contextual information in conversation is especially important to understand human emotions and intents \cite{DBLP:journals/corr/abs-2108-11626,DBLP:conf/icassp/HuHWJM22}. Based on this observation, we concatenate current utterance $u_{i}$ with its former 2-turn utterances $\{u_{i-1}, u_{i-2}\}$, and its latter 2-turn utterances $\{u_{i+1}, u_{i+2}\}$ as raw text. Additionally, we set segment id $S^{t}_{i}$ to distinguish utterance $u_{i}$ and its contexts in textual modality:
\begin{align}
\begin{split}
&I^t_{i}=[u_{i-2},u_{i-1},u_{i},u_{i+1},u_{i+2}]\\
&S^{t}_{i} = [\underbrace{0,\cdots,0}_{u_{i-2}, u_{i-1}},\underbrace{1,\cdots,1}_{u_{i}},\underbrace{0,\cdots,0}_{u_{i+1}, u_{i+2}}]
\end{split}
\end{align}
where the utterances are processed into the format of $I^{t}_{i}$, and we take $I^{t}_{i}$ as the textual modality of $I_{i}$.
Furthermore, we process raw acoustic input into numerical sequential vectors by librosa \footnote{https://github.com/librosa/librosa.} to extract Mel-spectrogram as audio features. It is the short-term power spectrum of sound and is widely used in modern audio processing. For video, we extract fixed T frames from each segment and use effecientNet \cite{DBLP:conf/icml/TanL19} pre-trained (supervised) on VGGface \footnote{https://www.robots.ox.ac.uk/~vgg/software/vgg\_face/.} and AFEW dataset to obtain video features.

\subsubsection{Label Formalization}
\begin{figure}[t]
\centerline{\includegraphics[width=0.9\linewidth]{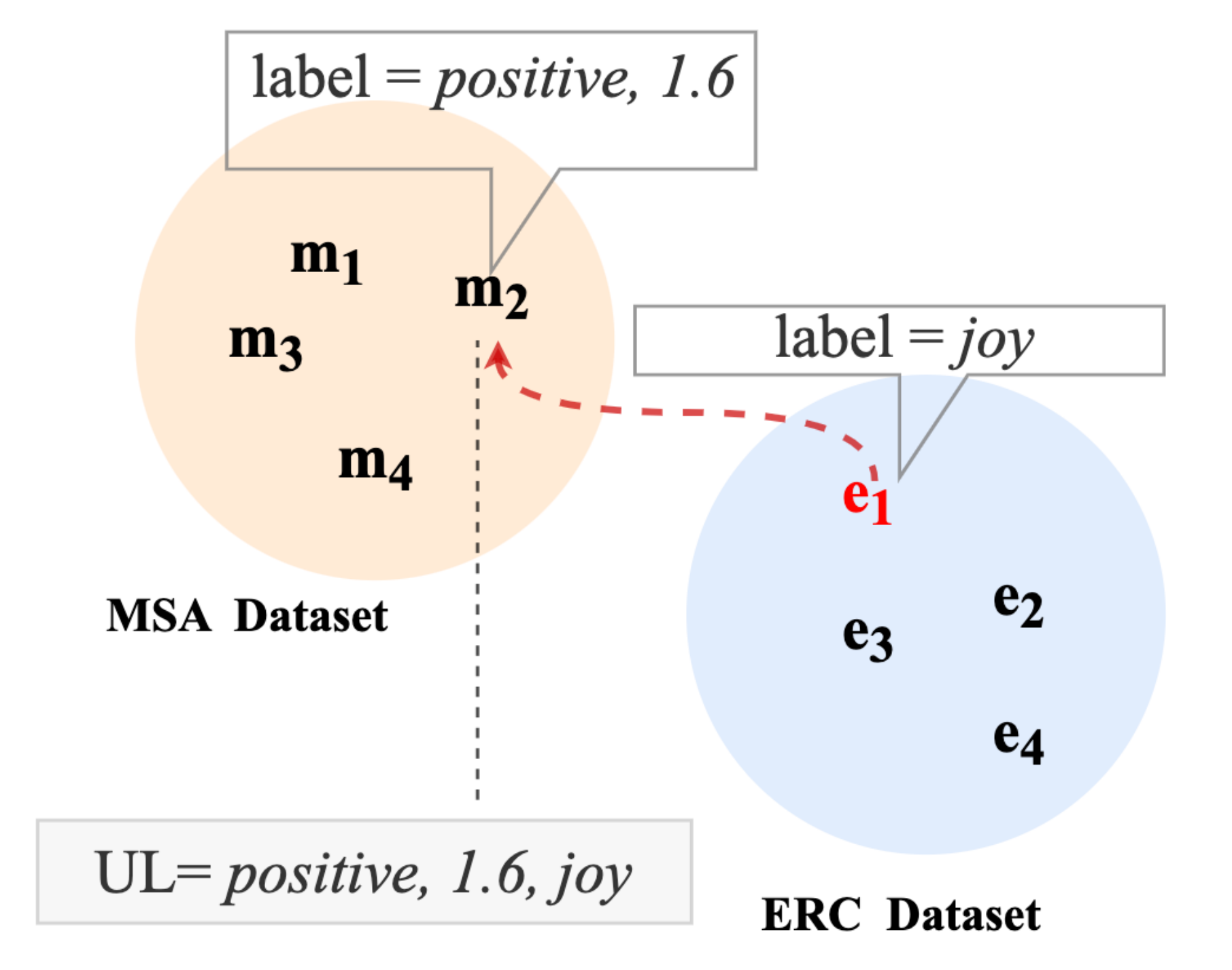}}
\caption{The generating process of a universal label (UL) and the red dashed line denotes that $e_{1}$ is the sample with the maximal semantic similarity to $m_{2}$.}
\label{fig:UL}
\end{figure}
To break the information boundary between MSA and ERC, we design a universal label (UL) scheme and take UL as the target sequence of UniMSE. The universal label aims to fully explore the shared knowledge between MSA and ERC on sentiment and emotion. Given a universal label $y_{i}=\{y_{i}^{p},y_{i}^{r},y_{i}^{c}\}$, it is composed by sentiment polarity $y^{p}_{i}\in \text{\{positive, negative and neutral\}}$ contained in MSA and ERC, sentiment intensity $y^{r}_{i}$ (the supervision signal of MSA task, a real number ranged from -3 to +3) and an emotion category $y^{c}_{i}$ (the supervision signal of ERC task, a predefined emotion category). 
We align the sample with similar semantics (like 1.6 and joy), in which one is annotated with sentiment intensity, and the other is annotated with emotion category. After the alignment of label space, each sample's label is formalized into a universal label format. Next, we introduce in detail how to unify MSA and ERC tasks in the label space, as follows:
% Universal label is 
% Pseudo-label (PL) has been successfully applied in many tasks, such as image classification \cite{DBLP:journals/kbs/MajumderHGCP18}, object detection \cite{DBLP:conf/iclr/GeCL20} and text classification \cite{DBLP:conf/nips/MukherjeeA20}. 

First, we classify the samples of MSA and ERC into positive, neutral, and negative sample sets according to their sentiment polarity. Then we calculate the similarity of two samples with the same sentiment polarity but belonging to different annotation scheme, thereby completing the missing part in the universal label. We show an example in Figure \ref{fig:UL}. Given an MSA sample $m_{2}$, it carries a positive sentiment and an annotation score of 1.6. Benchmarking the format of universal label, $m_{2}$ lacks an emotion category label. In this example, $e_{1}$ has the maximal semantic similarity to $m_{2}$, and then we assign the emotion category of $e_{1}$ as $m_{2}$'s emotion category. 

Previous works \cite{DBLP:conf/acl/TsaiBLKMS19,DBLP:conf/naacl/YangWYZRZPM21} have demonstrated that textual modality is more indicative than the other modalities, so we adopt textual similarity as the semantic similarity among samples. Specifically, we utilize a strong sentence embedding framework SimCSE \cite{DBLP:conf/emnlp/GaoYC21} to calculate the semantic similarity of two texts for completion of the universal label. Similarly, the sample in the ERC dataset is assigned a real number by calculating the MSA sample that is most similar to it. After our formalization, the samples of MSA and ERC are processed into $\{(I_0,y_0),(I_1,y_1)...(I_N,y_N)\}$, where $I_{i}$ denotes raw multimodal signal of sample $i$ and $y_{i}$ denotes its universal label. We can obtain the predictions of MSA and ERC by decoding from the predicted UL.
% For example, given a sample $s_{r}\in D_{r}$ and $D_{r}$ is the benchmark dataset of MSA, it carries a positive sentiment and an annotation score of 2.5. Benchmarking UL, $s_{r}$ lacks an emotion category label. We calculate the sample $s_{c}\in D_{c}$ that has the maximal similarity to $s_{r}$ in semantics by SimCSE \cite{DBLP:conf/emnlp/GaoYC21}, where $D_{c}$ is the benchmark dataset of ERC. After obtaining $s_{c}$, we assign the emotion category of $s_{c}$ as $s_{r}$'s emotional category. 
Additionally, we evaluate the performance of the generated automatically part in the universal labels. We randomly selected 80 samples with universal labels from MOSI and manually evaluated the generated labels used for universal label completion; the accuracy is about 90\%.

\subsection{Pre-trained Modality Fusion (PMF)}
%%% 介绍T5
Unlike the previous works just using a pre-trained model (such as T5) as a text encoder, we embed the multimodal fusion layers into the pre-trained model. Thus the acoustic and visual signals can participate in text encoding and are fused with multiple levels of textual information. The low-level text syntax feature encoded by the shallow Transformer layers and high-level text semantic feature encoded by the deep Transformer layers \cite{DBLP:conf/naacl/PetersNIGCLZ18,DBLP:conf/nips/VaswaniSPUJGKP17} are fused with audio and video features into multimodal representation. Besides, the injection of audio and vision to T5 can probe the relevant information in the massive pre-trained text knowledge, thereby incorporating richer pre-trained understanding into the multimodal fusion representation. We name this multimodal fusion process as pre-trained modality fusion (PMF).

We use T5 as the backbone of UniMSE. T5 contains multiple stacked Transformer layers, and each Transformer layer for the encoder and decoder contains a feedforward layer. The multimodal fusion layer is set to follow after the feedforward layer. Essentially, the PMF unit in the first Transformer layer of T5 receives a triplet $M_{i}=(X_{i}^{t}, X_{i}^{a}, X_{i}^{v})$ as the input, where $X_{i}^{m}, X_{i}^{m}\in R^{l_{m} \times d_{m}}$ denotes the modality representation of $I_{i}^{m}, m\in\{t,a,v\}$, $l_{m}$ and $d_m$ are the sequence length and the representation dimension of modality $m$, respectively. We view the multimodal fusion layer as an adapter \citep{DBLP:conf/icml/HoulsbyGJMLGAG19} and insert it into the T5 model to optimize specific parameters for multimodal fusion. The multimodal fusion layer receives modal representation triplet $M_{i}$, and maps the multimodal concatenation representation's size back to the layer’s input size. Specifically, we concatenate the three modal representations and then feed the concatenation into the down-projection and up-projection layers to fuse representations. For $j$-th PMF, the multimodal fusion is given by: 
\begin{align}
\begin{split}
&F_{i} = [F_{i}^{(j-1)}\cdot X_{i}^{a,l_{a}}\cdot X_{i}^{v,l_{v}}]\\
&F^{d}_{i} = \sigma(W^{d}F_{i} + b^{d})\\
&F^{u}_{i} = W^{u}F^{d}_{i} + b^{u}\\
&F^{(j)}_{i} = W(F_{i}^{u}\odot F_{i}^{(j-1)})
\end{split}
\end{align}
where $X_{i}^{a,l_{a}}\in R^{1\times d_a}$ and $X_{i}^{v,l_{v}}\in R^{1\times d_v}$ are hidden states of last time step of $X_{i}^{a}$ and $X_{i}^{v}$, respectively.
$X_{i}^{a}$ and $X_{i}^{v}$ are acoustic and visual modality representations encoded by two individual LSTMs, respectively.
$[\cdot]$ is concatenation operation on feature dim, $\sigma$ is the Sigmoid function, and $\{W^{d}, W^{u}, W, b^{d}, b^{u}\}$ are learnable parameters. $F_{i}^{(0)}=X_{i}^{t}$ and $X_{i}^{t}$ is the text representation encoded by the first Transformer layer of T5, and $F_{i}^{(j-1)}$ denotes the fusion representation after $(j-1)$ Transformer layers. $\odot$ denotes the element addition. The output of the fusion layer is then passed directly into the following layer normalization \cite{DBLP:journals/corr/BaKH16}.

Although we can embed a multimodal fusion layer in each Transformer of T5's encoder and decoder, it may bring two shortcomings: 1) disturb the encoding of text sequences, and 2) cause overfitting as more parameters are set for the multimodal fusion layer. Considering these issues, we use the former $j$ Transformer layers to encode the text, and the remaining Transformer layers are injected with the non-verbal (i.e., acoustic and visual) signals.

% \paragraph{Crossmodal Attention Multimodal Fusion}
% Cross-modal attention multimodal fusion layer focuses on the crossmodal interactions at the scale of the entire sequence. Given two modalities $u$ and $v$, their modal representations from each of them are with denoted $X_{u} \in R^{l_{u}\times d_{u}}$ and $X_{v} \in R^{l_{v}\times d_{v}}$, respectively, where $l(·)$ and $d(·)$ are used to represent sequence length and feature dimension, respectively. 
% We define the text unimodal as query to make the other two unimodals align to it. The alignment between unimodals refers to the crossmodal interactions at small granularity (i.e., token level for text and frame level for acoustic and visual). For unimodal pair $(X_{t}, X_{a})$, the crossmodal attention multimodal fusion is given by:
% \begin{align}
% \begin{split}
% &F_{i}^{(l\to a)}=Cross-Atten(T_{i}, A_{i}, A_{i})\\
% &\quad\quad\quad=softmax(\frac{Q^{T}_{i}K^{A}_{i}}{\sqrt{d}})V^{A}_{i}
% \end{split}
% \end{align}
% Similarly, for unimodal pair $(X_{t}, X_{v})$, the crossmodal attention multimodal fusion is given by:
% \begin{align}
% \begin{split}
% &F_{i}^{(l\to v)}=Cross-Atten(T_{i}, V_{i}, V_{i})\\
% &\quad\quad\quad=softmax(\frac{Q^{T}_{i}K^{V}_{i}}{\sqrt{d}})V^{V}_{i}
% \end{split}
% \end{align}

% \begin{align}
% &F_{i} = W[F_{i}^{(l\to a)}\cdot F_{i}^{(l\to v)}] + b
% \end{align}

\subsection{Inter-modality Contrastive Learning}
Contrastive learning (CL) has gained major advances in representation learning by viewing sample from multiple views \cite{DBLP:journals/jmlr/GutmannH10,DBLP:conf/nips/KhoslaTWSTIMLK20,DBLP:conf/emnlp/GaoYC21}. The principle of contrastive learning is that an anchor and its positive sample should be pulled closer, while the anchor and negative samples should be pushed apart in feature space. In our work, we perform inter-modality contrastive learning to enhance the interaction between modalities and magnify the differentiation of fusion representation among samples.
To ensure that each element of the input sequence is aware of its context, we process each modal representation to the same sequence length. We pass acoustic representation $X^{a}_{i}$, visual representations $X^{v}_{i}$ and fusion representation $F^{(j)}_{i}$ through a 1D temporal convolutional layer:
\begin{align}
\begin{split}
&\hat{X}^{u}_{i}=\text{Conv1D}(X^{u}_{i}, k^{u}),  u\in \{a,v\}\\
&\hat{F}^{(j)}_{i}=\text{Conv1D}(F^{(j)}_{i}, k^{f})
\end{split}
\end{align}
where $F^{(j)}_{i}$ is obtained after $j$ Transformer layers containing pre-trained modality fusion. $k^{u}$ is the size of the convolutional kernel for modalities $u,u\in \{a,v\}$, $k^{f}$ is the size of the convolutional kernel for fusion modality.
% \begin{align}
% \begin{split}
% &\hat{Z}^{m,j}_{i}=\text{Conv1D}(Z^{m,j}_{i}, k^{m})\in R^{d},  m\in \{t,a,v\}\\
% \end{split}
% \end{align}
% where $Z^{t,j}_{i}=F^{j}_{i}$, $Z^{a,*}_{i}=X^{a}_{i}$, $Z^{v,*}_{i}=X^{v}_{i}$ and $*$ denotes its value is irrelevant to $j$.

We construct each mini-batch with $K$ samples (each sample consists of acoustic, visual, and text modalities). Previous works \cite{DBLP:conf/emnlp/HanCP21,DBLP:conf/acl/TsaiBLKMS19} have proved that textual modality is more important than the other two modalities, so we take the textual modality as the anchor and the other two modalities as its augmented version.

A batch of randomly sampled pairs for each anchor consists of two positive pairs and $2K$ negative pairs. Here, the positive sample is the modality pair composed of text and corresponding acoustic in the same sample, and the modality pair composed of text and corresponding visual in the same sample. The negative example is the modality pair composed of text and the other two modalities of the other samples. For each anchor sample, the self-supervised contrastive loss is formulated as follows:
\begin{align}
\begin{split}
&L^{ta,j} = -\log \frac{\text{exp}(\hat{F}^{(j)}_{i}\hat{X}^{a}_{i})}{\text{exp}(\hat{F}^{(j)}_{i}\hat{X}^{a}_{i})+\sum_{k=1}^{K}\text{exp}(\hat{F}^{(j)}_{i}\hat{X}_{k}^{a})}\\
&L^{tv,j} = -\log \frac{\text{exp}(\hat{F}^{(j)}_{i}\hat{X}^{v}_{i})}{\text{exp}(\hat{F}^{(j)}_{i}\hat{X}^{v}_{i})+\sum_{k=1}^{K}\text{exp}(\hat{F}^{(j)}_{i}\hat{X}_{k}^{v})}
\end{split}
\end{align}
where $L^{ta,j}$ and $L^{tv,j}$ represent the contrastive loss of text-acoustic and text-visual performing on the $j$-th Transformer layer of encoder, respectively.

\subsection{Grounding UL to MSA and ERC}
During the training phase, we use the negative log-likelihood to optimize the model, which takes the universal label as the target sequence. The overall loss function can be formulated as follows:
\begin{align}
L = L^{task} + \alpha (\sum\nolimits_{j}L^{ta,j}) + \beta (\sum\nolimits_{j}L^{tv,j})
\end{align}
where $L^{task}$ denotes the generative task loss, $j$ is the index of the Transformer layer of the Encoder, and $\{\alpha, \beta\}$ are decimals between 0 and 1, indicating the weight values. Moreover, during the inference, we use the decoding algorithm \footnote{Please see Appendix \ref{sec:dccoding} for details.} to convert the output sequence into the real number for MSA and the emotion category for ERC.

\section{Experiments}
\subsection{Datasets}
We conduct experiments on four publicly available benchmark datasets of MSA and ERC, including Multimodal Opinion-level Sentiment Intensity dataset (MOSI) \cite{DBLP:journals/expert/ZadehZPM16}, Multimodal Opinion Sentiment and Emotion Intensity (MOSEI) \cite{DBLP:conf/acl/MorencyCPLZ18}, Multimodal EmotionLines Dataset (MELD) \cite{DBLP:conf/acl/PoriaHMNCM19} and Interactive Emotional dyadic Motion CAPture database (IEMOCAP) \cite{DBLP:journals/lre/BussoBLKMKCLN08}. The detailed statistic of four datasets are shown in Table \ref{tab:data}. More details can see Appendix \ref{sec:duration}.

\textbf{MOSI} contains 2199 utterance video segments, and each segment is manually annotated with a sentiment score ranging from -3 to +3 to indicate the sentiment polarity and relative sentiment strength of the segment. \textbf{MOSEI} is an upgraded version of MOSI, annotated with sentiment and emotion. MOSEI contains 22,856 movie review clips from YouTube. Most existing studies only use MOSEI’s sentiment annotation, and MOSEI’s emotion annotation is multiple labels, so we do not use 
its emotion annotation although they are available. Note that there is no overlap between MOSI and MOSEI, and the data collection and labeling processes for the two datasets are independent.

% and its annotation style is the same as MOSI. 

\textbf{IEMOCAP} consists of 7532 samples. Following previous works  \cite{DBLP:conf/aaai/WangSLLZM19,DBLP:conf/icassp/HuHWJM22}, we select six emotions for emotion recognition, including joy, sadness, angry, neutral, excited, and frustrated. \textbf{MELD} contains 13,707 video clips of multi-party conversations, with labels following Ekman’s six universal emotions, including joy, sadness, fear, anger, surprise and disgust.
% We follow previous works \cite{DBLP:conf/emnlp/Mao0WGL21,DBLP:conf/icassp/HuHWJM22} and adopt the setting.
% are generally considered as annotation labels in the previous works \cite{DBLP:conf/emnlp/Mao0WGL21,DBLP:conf/icassp/HuHWJM22}.

\subsection{Evaluation metrics}
For MOSI and MOSEI, we follow previous works \cite{DBLP:conf/emnlp/HanCP21} and adopt mean absolute error (MAE), Pearson correlation (Corr), seven-class classification accuracy (ACC-7), binary classification accuracy (ACC-2) and F1 score computed for positive/negative and non-negative/negative classification as evaluation metrics. For MELD and IEMOCAP, we use accuracy (ACC) and weighted F1 (WF1) for evaluation.
% (i.e., ACC-7: sentiment score classification),

\begin{table}[]
\centering
\resizebox{0.49\textwidth}{!}{\begin{tabular}{lcccccc}
\toprule
          & \textbf{Train} & \textbf{Valid} & \textbf{Test} & \textbf{All}& \textbf{Senti.} & \textbf{Emo.} \\
\midrule
MOSI  & 1284           & 229                 & 686       &  2199  &  \Checkmark                         &    \XSolidBrush                         \\
MOSEI & 16326          & 1871                & 4659    &  22856   &  \Checkmark                         &    \Checkmark                         \\
MELD      & 9989    & 1108         & 2610    &   13707      &  \XSolidBrush                         &     \Checkmark                       \\
IEMOCAP   & 5354    & 528           & 1650   & 7532                 &  \XSolidBrush                         &     \Checkmark                        \\
\bottomrule
\end{tabular}}
\caption{The details of MOSI, MOSEI, MELD, and IEMOCAP, including data splitting and the labels it contains, where Senti. and Emo. represent the label sentiment polarity and intensity of MSA and emotion category of ERC, respectively. \Checkmark and \XSolidBrush denote the dataset has or does not have the label.}
\label{tab:data}
\end{table}

\begin{table*}[]
\resizebox{\textwidth}{!}{
\begin{tabular}{c|ccccc|ccccc|cc|cc}
\toprule
 & \multicolumn{5}{c}{MOSI} & \multicolumn{5}{c}{MOSEI} & \multicolumn{2}{c}{MELD} & \multicolumn{2}{c}{IEMOCAP} \\
\toprule
\textbf{Method} & \textbf{MAE}$\downarrow$ & \textbf{Corr}$\uparrow$ & \textbf{ACC-7}$\uparrow$ & \textbf{ACC-2}$\uparrow$ & \textbf{F1}$\uparrow$ & \textbf{MAE}$\downarrow$ & \textbf{Corr}$\uparrow$ & \textbf{ACC-7}$\uparrow$ & \textbf{ACC-2}$\uparrow$ & \textbf{F1}$\uparrow$ & \textbf{ACC}$\uparrow$ & \textbf{WF1}$\uparrow$ & \textbf{ACC}$\uparrow$ & \textbf{WF1}$\uparrow$ \\
\midrule
LMF & 0.917 & 0.695 & 33.20 & -/82.5 & -/82.4 & 0.623 & 0.700 & 48.00 & -/82.0 & -/82.1 & 61.15 & 58.30 & 56.50 & 56.49 \\
TFN & 0.901 & 0.698 & 34.90 & -/80.8 & -/80.7 & 0.593 & 0.677 & 50.20 & -/82.5 & -/82.1 & 60.70 & 57.74 & 55.02 & 55.13 \\
MFM & 0.877 & 0.706 & 35.40 & -/81.7 & -/81.6 & 0.568 & 0.703 & 51.30 & -/84.4 & -/84.3 & 60.80 & 57.80 & 61.24 & 61.60 \\
MTAG & 0.866 & 0.722 & 38.90 & -/82.3 & -/82.1 & - & - & - & - & - & - & - & - & - \\
SPC & - & - & - & -/82.8 & -/82.9 & - & - & - & -/82.6 & -/82.8 & - & - & - & - \\
ICCN & 0.862 & 0.714 & 39.00 & -/83.0 & -/83.0 & 0.565 & 0.704 & 51.60 & -/84.2 & -/84.2 & - & - & 64.00 & 63.50 \\
MulT & 0.861 & 0.711 & - & 81.50/84.10 & 80.60/83.90 & 0.580 & 0.713 & - & -/82.5 & -/82.3 & - & - & - & - \\
MISA & 0.804 & 0.764 & - & 80.79/82.10 & 80.77/82.03 & 0.568 & 0.717 & - & 82.59/84.23 & 82.67/83.97 & - & - & - & - \\
COGMEN & - & - & 43.90 & -/84.34 & - & - & - & - & - & - & - & - & 68.20 & 67.63 \\
Self-MM & 0.713 & 0.798 & - & 84.00/85.98 & 84.42/85.95 & 0.530 & 0.765 & - & 82.81/85.17 & 82.53/85.30 & - & - & - & - \\
MAG-BERT & 0.712 & 0.796 & - & {\ul 84.20/86.10} & {\ul 84.10/86.00} & - & - & - & {\ul 84.70/-} & {\ul 84.50/-} & - & - & - & - \\
MMIM & {\ul 0.700} & {\ul 0.800} & {\ul 46.65} & 84.14/86.06 & 84.00/85.98 & {\ul 0.526} & {\ul 0.772} & {\ul 54.24} & 82.24/85.97 & 82.66/85.94 & - & - & - & - \\
\textit{DialogueGCN} & - & - & - & - & - & - & - & - & - & - & 59.46 & 58.10 & 65.25 & 64.18 \\
\textit{DialogueCRN} & - & - & - & - & - & - & - & - & - & - & 60.73 & 58.39 & 66.05 & 66.20 \\
\textit{DAG-ERC} & - & - & - & - & - & - & - & - & - & - & - & 63.65 & - & 68.03 \\
\textit{ERMC-DisGCN} & - & - & - & - & - & - & - & - & - & - & - & 64.22 & - & 64.10 \\
\textit{CoG-BART*} & - & - & - & - & - & - & - & - & - & - & - & 64.81 & - & 66.18 \\
\textit{Psychological} & - & - & - & - & - & - & - & - & - & - & - & 65.18 & - & 66.96 \\
\textit{COSMIC} & - & - & - & - & - & - & - & - & - & - & - & 65.21 & - & 65.28 \\
\textit{TODKAT*} & - & - & - & - & - & - & - & - & - & - & - & {\ul 65.47} & - & 61.33 \\
MMGCN & - & - & - & - & - & - & - & - & - & - & - & 58.65 & - & 66.22 \\
MM-DFN & - & - & - & - & - & - & - & - & - & - & {\ul 62.49} & 59.46 & {\ul 68.21} & {\ul 68.18} \\
\textbf{UniMSE} & \textbf{0.691} & \textbf{0.809} & \textbf{48.68} & \textbf{85.85/86.9} & \textbf{85.83/86.42} & \textbf{0.523} & \textbf{0.773} & \textbf{54.39} & \textbf{85.86/87.50} & \textbf{85.79/87.46} & \textbf{65.09} & \textbf{65.51} & \textbf{70.56} & \textbf{70.66} \\
\bottomrule
\end{tabular} 
}
\caption{Results on MOSI, MOSEI, MELD, and IEMOCAP. *The performances of baselines are updated by their authors in the official code repository, and the baselines with italics indicate it only uses textual modality. The results with underline denote the previous SOTA performance.}
\label{tab:main_results}
\end{table*}

\subsection{Baselines}
% We perform comparative study against UniMSE by considering all baselines
We compare the proposed method with competitive baselines in MSA and ERC tasks. For MSA, the baselines can be grouped into 1) early multimodal fusion methods like {Tensor Fusion Network \bf TFN} \cite{DBLP:conf/emnlp/ZadehCPCM17}, {Low-rank Multimodal Fusion \bf LMF} \cite{DBLP:conf/acl/MorencyLZLSL18}, and {Multimodal Factorization Model \bf MFM} \cite{DBLP:conf/iclr/TsaiLZMS19}, and 2) the methods that fuse multimodality through modeling modality interaction, such as multimodal Transformer {\bf MulT} \cite{DBLP:conf/acl/TsaiBLKMS19}, interaction canonical correlation network {\bf ICCN} \cite{DBLP:conf/aaai/SunSSL20}, sparse phased Transformer {\bf SPC} \cite{DBLP:conf/emnlp/ChengFB021}, and modal-temporal attention graph {\bf MTAG} \cite{DBLP:conf/naacl/YangWYZRZPM21} and 3) the methods focusing on the consistency and the difference of modality, in which {\bf MISA} \cite{DBLP:conf/mm/HazarikaZP20} controls the modal representation space, {\bf Self-MM} \cite{DBLP:conf/aaai/YuXYW21} learns from unimodal representation using multi-task learning, {\bf MAG-BERT} \cite{DBLP:conf/acl/RahmanHLZMMH20} designs a fusion gate, and {\bf MMIM} \cite{DBLP:conf/emnlp/HanCP21} hierarchically maximizes the mutual information. 

With the rise of multimodal information, {\bf MMGCN} \cite{DBLP:conf/acl/HuLZJ20}, {\bf MM-DFN} \cite{DBLP:conf/icassp/HuHWJM22} and {\bf COGMEN} \cite{DBLP:journals/corr/abs-2205-02455} consider the multimodal conversational context to solve ERC task. Some works only use textual modality to recognize emotion, in which {\bf ERMC-DisGCN} \cite{DBLP:conf/emnlp/SunYF21}, 
{\bf Psychological} \cite{DBLP:conf/emnlp/Li00W21}, {\bf DAG-ERC} \cite{DBLP:conf/acl/ShenWYQ20} and {\bf DialogueGCN} \cite{DBLP:conf/emnlp/GhosalMPCG19} adapt the GNN-based model to capture contexts. Additionally, {\bf CoG-BART} \cite{DBLP:journals/corr/abs-2112-11202} learns the context knowledge from the pre-trained model, {\bf COSMIC} \cite{DBLP:conf/emnlp/GhosalMGMP20} incorporates different elements of commonsense, and {\bf TODKAT} \cite{DBLP:conf/acl/ZhuP0ZH20} uses topic-driven knowledge-aware Transformer to model affective states.
Similar to MSA and ERC works, UniMSE still attends to improve multimodal fusion representation and modality comparison in feature space. But, UniMSE unifies MSA and ERC tasks into a single architecture to implement knowledge-sharing.

\subsection{Experimental Settings} % Hyperparameter Setting
We use pre-trained T5-Base \footnote{https://github.com/huggingface/transformers/tree/main\\/src/transformers/models/t5.} as the backbone of UniMSE. We integrate the training sets of {MOSI, MOSEI, MELD, IEMOCAP} to train the model and valid sets to select hyperparameters. The batch size is 96, the learning rate for T5 fine-tuning is set at 3e-4, and the learning rate for main and pre-trained modality fusion are 0.0001 and 0.0001, respectively. The hidden dim of acoustic and visual representation is 64, the T5 embedding size is 768, and the fusion vector size is 768. We insert a pre-trained modality fusion layer into the last 3 Transformer layers of T5's encoder. The contrastive learning performs the last 3 Transformer layers of T5's encoder, and we set $\alpha=0.5$ and $\beta=0.5$. More details can see Appendix \ref{sec:environment}.

\subsection{Results}

We compare UniMSE with the baselines on datasets MOSI, MOSEI, IEMOCAP, and MELD, and the comparative results are shown in Table \ref{tab:main_results}.
UniMSE significantly outperforms SOTA in all metrics on MOSI, MOSEI, IEMOCAP, and MELD. Compared to the previous SOTA, UniMSE improves ACC-2 of MOSI, ACC-2 of MOSEI, ACC of MELD, and ACC of IEMOCAP by 1.65\%, 1.16\%, 2.6\%, and 2.35\% respectively, and improves F1 of MOSI, F1 of MOSEI, and WF1 of IEMOCAP by 1.73\%, 1.29\%, and 2.48\% respectively.
It can be observed that early works like LMF, TFN, and MFM performed on the four datasets. However, the later works, whether MSA or ERC, only evaluate their models on partial datasets or metrics, yet we provide results on all datasets and corresponding metrics. For example, MTAG only conducts experiments on MOSI, and most ERC works only give the WF1, which makes MSA and ERC tasks tend to be isolated in sentiment knowledge. Unlike these works, UniMSE unifies MSA and ERC tasks on these four datasets and evaluates them based on the common metrics of the two tasks. In summary, 1) UniMSE performs on all benchmark datasets of MSA and ECR; 2) UniMSE significantly outperforms SOTA in most cases. These results illustrate the superiority of UniMSE in MSA and ERC tasks and demonstrate the effectiveness of a unified framework in knowledge sharing among tasks and datasets.

% \begin{table}[t]
%     \small
%   \begin{center}
%     \begin{tabular}{cccccccccc}\\
%     \hline
%   \multirow{2}*{}&
%     \multicolumn{2}{c}{\bf MELD}&\multicolumn{2}{c}{\bf IEMOCAP}\cr\cline{2-5}
%     &Acc&WF&Acc&WF\cr
%     \hline
%     TFN&60.7&57.74&55.02&55.13\\
%     LMF&61.15&58.30&56.50&56.49\\
%     MFN&60.80&57.80&61.24&61.60\\
%     bc-LSTM&59.62&57.29&60.51&60.42\\
%     ICON&-&-&64.00&63.50\\
%     DialogueRNN&60.31&57.66&63.52&62.89\\
%     DialogueCRN&61.11&58.67&67.16&67.21\\
%     DialogueGCN&60.42&58.31&63.22&62.89\\
%     MMGCN&60.42&58.31&66.36&66.26\\
%     MM-DFN&62.49&59.46&68.21&68.18\\
%     DialogueTRM-M&65.7&63.5&69.5&69.7\\
%     EmoBERTa&-&65.61&-&68.57\\
%     EmotionFlow-&65.05&-&-\\
%     HCL&-&68.73&-&66.18\\
%     CoMPM&-&65.77&-&69.46\\
%     \hline
%     \end{tabular}
%     \end{center}
%     \caption{Results on MELD and IEMOCAP.}
%     \label{tab:results_meld_iemocap}
% \end{table}
\subsection{Ablation Study}
We conducted a series of ablation studies on MOSI, and the results are shown in Table \ref{tab:ablation}. First, we eliminate one or several modalities from multimodal signals to verify the modal effects on model performance. We can find that removing visual and acoustic modalities or one of them all leads to performance degradation, which indicates that the non-verbal signals (i.e., visual and acoustic) are necessary for solving MSA, and demonstrates the complementarity among text, acoustic, and visual. We also find that the acoustic modality is more important than the visual to UniMSE. Then we eliminate module PMF and CL from UniMSE, which leads to an increase in MAE and a decrease in Corr. These results illustrate the effectiveness of PMF and CL in multimodal representation learning. 
Additionally, we conduct experiments to verify the impact of the dataset on UniMSE. We remove IEMOCAP, MELD, and MOSEI from the training set and evaluate model performance on the MOSI test set. Removing IEMOCAP and MELD hurts the performance, especially in metrics MAE and Corr. This result may be because the removal of MELD/IEMOCAP has reduced the information they provide for MSA task. We also remove MOSEI, resulting in poor performance in the four metrics. The proposed UniMSE is orthogonal to the existing works, and it is believed that introducing our unified framework to other tasks can also bring improvements.

% \begin{table}[t]
% \resizebox{0.49\textwidth}{!}{
% \begin{tabular}{clcccc}
% \toprule
%                           &        & \textbf{MAE}   & \textbf{Corr}  & \textbf{ACC-2}       & \textbf{F1}          \\
% \midrule
% \multirow{3}{*}{\rotatebox{90}{Modality}} & T     & 0.721 & 0.780 & 83.72/85.11 & 83.52/85.11 \\
%                           & T+A   & 0.714 & 0.798 & 84.37/85.37 & 84.71/85.78 \\
%                           & T+V   & 0.719 & 0.794 & 83.82/85.20 & 83.86/85.69 \\
% \hline
% \multirow{3}{*}{\rotatebox{90}{Module}}   & w/o PMF   & 0.722                & 0.785                & 85.13/86.59          & 85.03/86.37          \\
%                           & w/o CL    & 0.713                & 0.795                & 85.28/86.59          & 85.27/86.55          \\
%                           & & & & & \\

% \hline
%                           & UniMSE & 0.691 & 0.809 & 85.96/86.89 & 85.85/86.90 \\
% \bottomrule
% \end{tabular}}
% \caption{Ablation study of UniMSE on CMU-MOSI. T, V, A represent text, visual, acoustic respectively. PMF and CL represent pre-trained multimodal fusion and contrastive learning respectively.}
% \label{tab:ablation}
% \end{table}

\begin{table}[t]
\resizebox{0.49\textwidth}{!}{
\begin{tabular}{l|cccc}
\toprule
 & \textbf{MAE} & \textbf{Corr} & \textbf{ACC-2} & \textbf{F1} \\
\midrule
\textbf{UniMSE} & \textbf{0.691} & \textbf{0.809} & \textbf{85.85/86.9} & \textbf{85.83/86.42} \\
\midrule
- w/o A & 0.719 & 0.794 & 83.82/85.20 & 83.86/85.69 \\
- w/o V & 0.714 & 0.798 & 84.37/85.37 & 84.71/85.78 \\
- w/o A, V & 0.721 & 0.780 & 83.72/85.11 & 83.52/85.11 \\
\midrule
- w/o PMF & 0.722 & 0.785 & 85.13/86.59 & 85.03/86.37 \\
- w/o CL & 0.713 & 0.795 & 85.28/86.59 & 85.27/86.55 \\
\midrule
- w/o IEMOCAP & 0.718 & 0.784 & 84.11/85.88 &84.75/85.47 \\
- w/o MELD	& 0.722 & 0.776 & 84.05/84.96 &84.50/84.64 \\
- w/o MOSEI	&0.775&0.727&80.68/81.22&81.35/81.83 \\
\bottomrule
\end{tabular}}
\caption{Ablation study of UniMSE on MOSI. V and A represent visual and acoustic modalities, respectively. PMF and CL represent pre-trained modality fusion and contrastive learning, respectively.}
\label{tab:ablation}
\end{table}

\subsection{Visualization}
\begin{figure}[t]
\centering
\subfigure[\label{fig:visualization:a}]{
\includegraphics[width=0.45\linewidth]{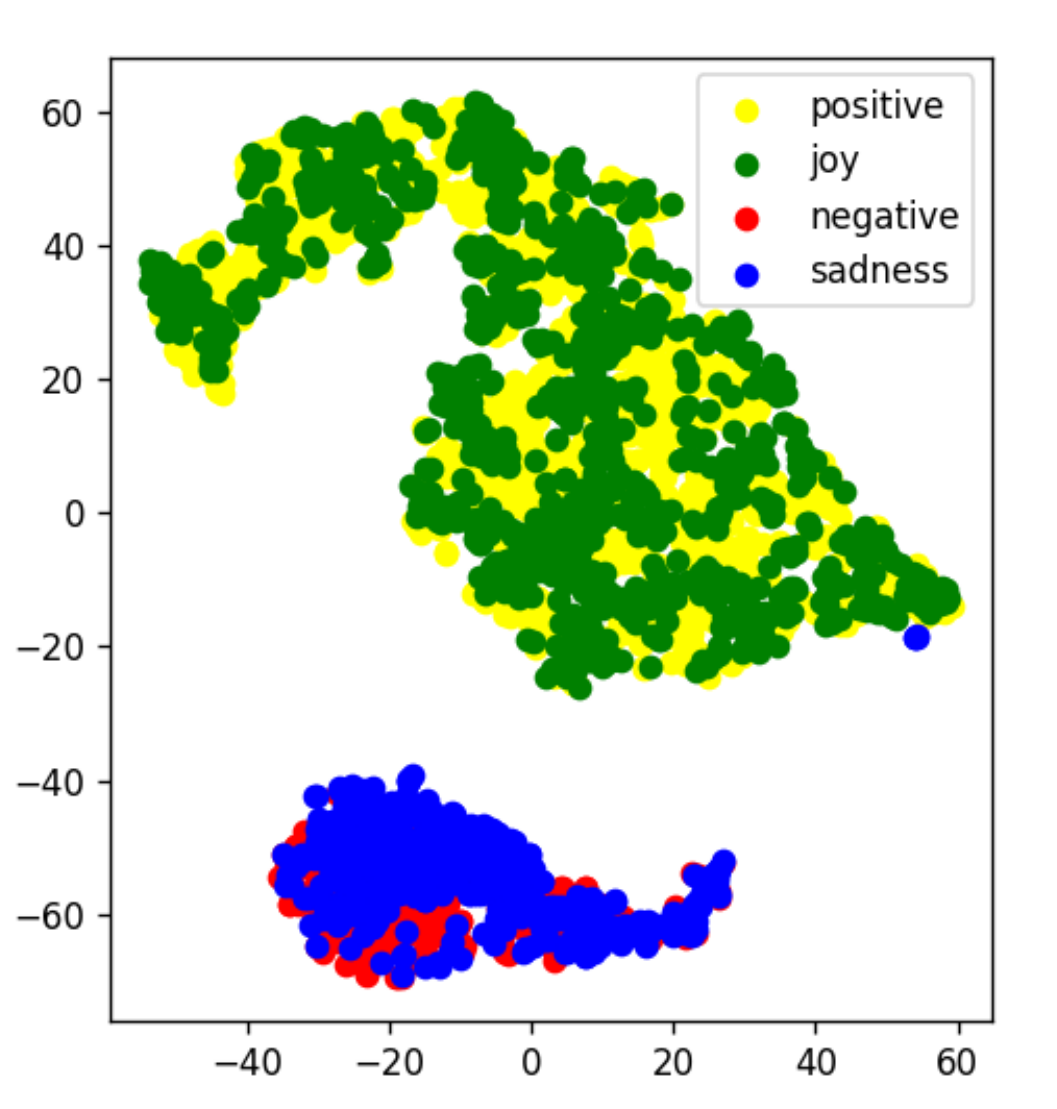}}
\hspace{0.05in}
\subfigure[\label{fig:visualization:b}]{
\includegraphics[width=0.45\linewidth]{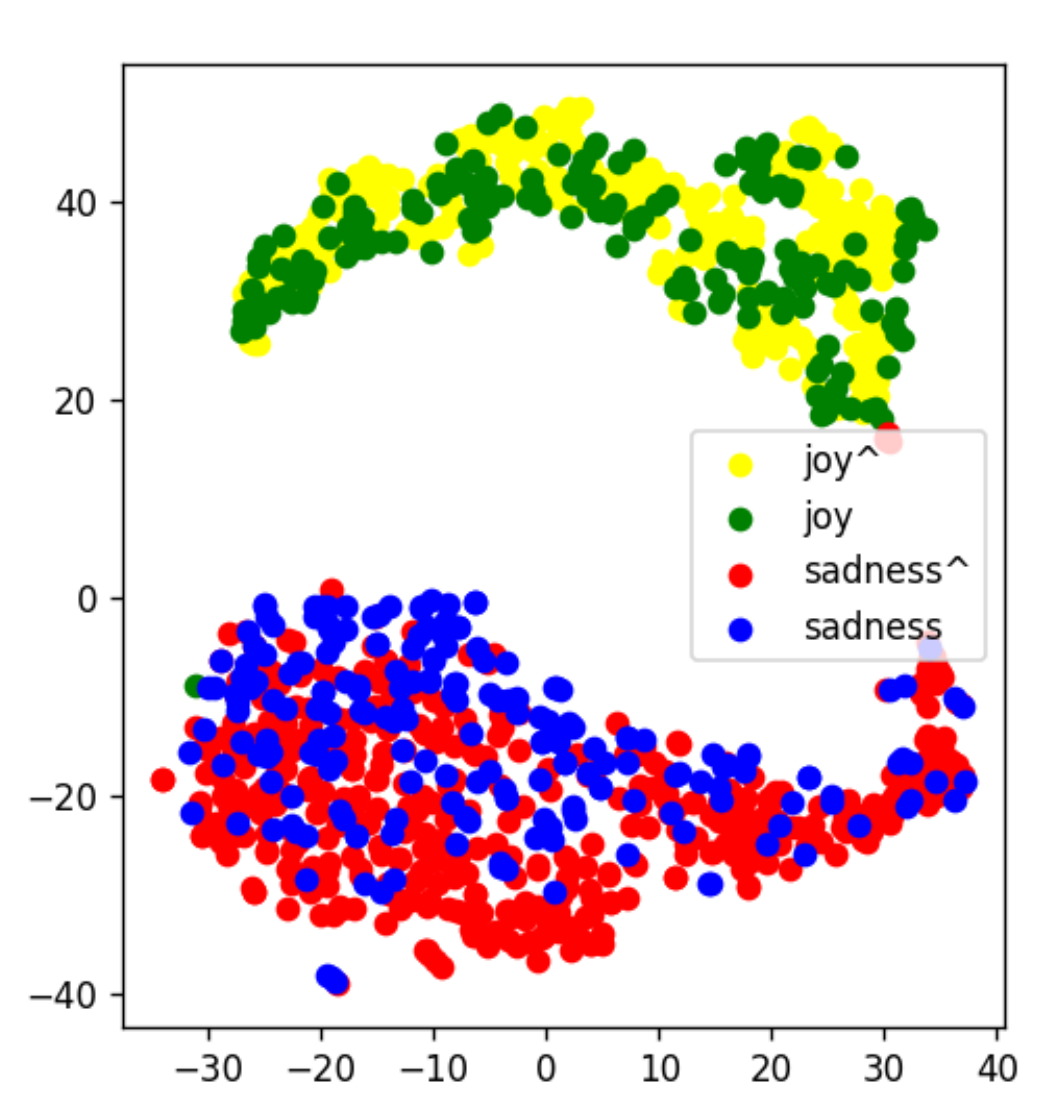}}
\caption{T-SNE visualization comparison of the multimodal fusion representation between: (a) samples with sentiment and emotion, and (b) samples with original emotion and generated emotion, where joy\^ ~ and sadness\^ ~ denote the generated emotion.}
\label{fig:visualization}
\end{figure}

To verify the effects of UniMSE's UL and cross-task learning on multimodal representation, we visualize multimodal fusion representation (i.e., $F_{i}^{(j)}$) of the last Transformer layer. Specifically, we select samples that carry positive/negative sentiment polarity from the test set of MOSI and select samples that have the joy/sadness emotion from the test set of MELD. Their representation visualization is shown in Figure \ref{fig:visualization:a}. It can be observed that the representations of samples with positive sentiment cover the representation of samples with joy emotion, which demonstrates that although these samples are from different tasks, a common feature space exists between the samples with joy emotion and positive sentiment.

Moreover, we also select the MOSI samples with generated emotion joy/sadness and compare them to MELD samples with the original emotion label joy/sadness in embedding space. Their visualization is shown in Figure \ref{fig:visualization:b}. The samples with joy emotion, whether annotated with the original label or generated based on UL, share a common feature space. These results verify the superiority of UniMSE on representation learning across samples and demonstrate the complementarity between sentiment and emotion.

% model parameters and loss function across MSA and ERC
% We believe this work can inspire future creativity in representation learning and multimodal sentiment/emotion analysis.

\section{Conclusion}
This paper provides a psychological perspective to demonstrate that jointly modeling sentiment and emotion is feasible and reasonable. We present a unified multimodal knowledge-sharing framework, UniMSE, to solve MSA and ERC tasks. UniMSE not only captures knowledge of sentiment and emotion, but also aligns the input features and output labels. Moreover, we fuse acoustic and visual modal representation with multi-level textual features and introduce inter-modality contrastive learning. We conduct extensive experiments on four benchmark datasets and achieve SOTA results in all metrics. We also provide the visualization of multimodal representation, proving the relevance of sentiment and emotion in embedding space. We believe this work presents a new experimental setting that can provide a new and different perspective to the MSA and ERC research communities.

% \subsection{Appendices}

\section*{Limitations}
%%% 在整合上下文时， 我们只在ERC上iemocap和meld考虑了上下文信息，并且需要下文
In this preliminary study, we only integrate context information on MELD and IEMOCAP, and the context on MOSI and MOSEI will be considered in the future. Furthermore, the generation of universal labels only considers textual modality, without considering acoustic and visual modalities, which will also be solved in our future work. 
% Furthermore, the latter two utterances of current utterance in the conversation are viewed its context in our work, which may lead to the risk of information leakage  when identifying the emotions of the current utterance, which will also be solved in our future work. 

\section*{Ethics Statement}
The data used in this study are all open-source data for research purposes. While making machines understand human emotions sounds appealing, it could be applied to emotional companion robots or intelligent customer service. However, even in simple six-class emotion recognition (MELD), the proposed method can achieve only 65\% in accuracy, which is far from usable.

\section*{Acknowledgement}
This work was supported by the Natural Science Foundation of Guangdong, China under Grant No. 2020A1515010812 and
2021A1515011594.

\bibliography{emnlp2022}
\bibliographystyle{acl_natbib}
\newpage
\appendix

\section{Appendix}
\subsection{Datasets}
\label{sec:duration}
We count the duration of the video segment in MSA and ERC and give the results in Table \ref{tab:duration}. We take the length of the video segment as the duration of sentiment or emotion. We can observe that the average time of sentiment in MSA is longer than that of emotion in ERC, demonstrating the difference between sentiment and emotion. 
The average length of the video segment in MOSEI is 7.6 seconds. This may indicate why MOSEI is usually used to study sentiments rather than emotions. Furthermore, we count emotion categories of MELD and IEMOCAP, and their distributions of the train set, valid set, and test set are shown in Table \ref{tab:meld_dist} and Table \ref{tab:iemocap_dist}, respectively.
\begin{table}[h]
\begin{tabular}{c|ccc}
\toprule
\textbf{Task}        & \textbf{Dataset} & \textbf{D-A VL (s)} & \textbf{T-A VL (s)} \\
\midrule
\multirow{2}{*}{MSA} & MOSI             & 4.2                                             & \multirow{2}{*}{7.3}                         \\
                     & MOSEI            & 7.6                                             &                                              \\
\midrule
\multirow{2}{*}{ERC} & MELD             & 3.2                                             & \multirow{2}{*}{3.7}                         \\
                     & IEMOCAP          & 4.6                                             &       \\
\bottomrule
\end{tabular}
\caption{Average video length of samples. D-A VL(s) and T-A VL(s) denote the average video length of datasets and tasks, respectively.}
\label{tab:duration}
\end{table}
\begin{table}[h]
\resizebox{0.49\textwidth}{!}{
\begin{tabular}{c|cccccc}
\toprule
      & \textbf{surprise} & \textbf{fear} & \textbf{sadness} & \textbf{joy}  & \textbf{disgust} & \textbf{anger} \\
      \midrule
train & 1205     & 268  & 683     & 1744 & 271     & 1109  \\
dev   & 150      & 40   & 112     & 163  & 22      & 153   \\
test  & 281      & 50   & 208     & 402  & 68      & 345   \\
all   & 1636     & 358  & 1003    & 2309 & 361     & 1607 \\
\bottomrule
\end{tabular}}
\caption{The distribution of emotion category on dataset MELD.}
\label{tab:meld_dist}
\end{table}
\begin{table}[t]
\resizebox{0.49\textwidth}{!}{
\begin{tabular}{c|cccccc}
\toprule
      & \textbf{neural} & \textbf{frustrated} & \textbf{angry} & \textbf{sadness} & \textbf{joy} & \textbf{excited} \\
\midrule
train & 1187   & 1322       & 832   & 762     & 431   & 703     \\
dev   & 137    & 146        & 101   & 77      & 21    & 39      \\
test  & 384    & 381        & 170   & 245     & 299   & 143     \\
all   & 1708   & 1849       & 1103  & 1084     & 751   & 885    \\
\bottomrule
\end{tabular}}
\caption{The distribution of emotion category on dataset IEMOCAP.}
\label{tab:iemocap_dist}
\end{table}

\begin{algorithm}[h]
\caption{Decoding Algorithm for MSA and ERC tasks}%算法名字
\label{decoding_algorithm}
\LinesNumbered %要求显示行号
\KwIn{Target task $t\in \{MSA, ERC\}$, target sequence $Y=\{y_{1},y_{2},\cdots,y_{N}\}$ and $y_{i}=(y_{i}^{p}, y_{i}^{r}, y_{i}^{c})$}%输入参数
\KwOut{Task prediction $Y^{t}=\{y_{1}^{t},y_{2}^{t},\cdots,y_{N}^{t}\}$ for target $t$}%输出
  ~~$Y^{t}=\{\}$\\
  \For{each $y_{i}$ in $Y$}{
      ~~$y_{i}^{r}$=$y_{i}$[1]\\
      ~~$y_{i}^{c}$=$y_{i}$[2]\\
      \If{t is MSA}{
        $Y^{t}$.append($y_{i}^{r}$)\\
      }
      \If{t is ERC}{
        $Y^{t}$.append($y_{i}^{c}$)\\
      }
    }
  return $Y^{t}$
\end{algorithm}

\subsection{Decoding Algorithm for MSA and ERC tasks}
\label{sec:dccoding}
In this part, we introduce the decoding algorithm we used to convert the predicted target sequence of UniMSE into a sentiment intensity for MSA and an emotion category for ERC. The algorithm is shown in Algorithm \ref{decoding_algorithm}.

\subsection{Experimental Environment}
\label{sec:environment}
All experiments are conducted in the NVIDIA RTX A100 and NVIDIA RTX V100. The T5-base model has 220M parameters, including 12 layers, 768 hidden dimensions, and 12 heads. PML contains two projection layers, and its parameter number is nearly a thousandth of the original parameter number.

\end{document}